\documentclass[10pt,twocolumn,letterpaper]{article}

\usepackage[pagenumbers]{cvpr} 

\usepackage[dvipsnames]{xcolor}

\definecolor{cvprblue}{rgb}{0.21,0.49,0.74}
\usepackage[pagebackref,breaklinks,colorlinks,citecolor=cvprblue]{hyperref}

\usepackage{algorithm}
\usepackage[noend]{algpseudocode}
\usepackage{graphicx}
\usepackage{amsmath}
\usepackage{amssymb}
\usepackage{booktabs}
\usepackage{multirow}
\usepackage{amssymb}
\usepackage{fontawesome}

\def\paperID{21} 
\def\confName{CVPR}
\def\confYear{2024}

\title{MultIOD: Rehearsal-free Multihead Incremental Object Detector}

\author{Eden Belouadah ~~~~ Arnaud Dapogny ~~~~ Kevin Bailly\\
Datakalab, 143 avenue Charles de Gaulles, 92200 Neuilly-sur-Seine, France\\
{\tt\small \{eb, ad, kb\}@datakalab.com}
}

\begin{document}
\maketitle
\begin{abstract}
Class-Incremental learning (CIL) refers to the ability of artificial agents to integrate new classes as they appear in a stream. It is particularly interesting in evolving environments where agents have limited access to memory and computational resources. The main challenge of incremental learning is catastrophic forgetting, the inability of neural networks to retain past knowledge when learning a new one. Unfortunately, most existing class-incremental methods for object detection are applied to two-stage algorithms such as Faster-RCNN, and rely on rehearsal memory to retain past knowledge. We argue that those are not suitable in resource-limited environments, and more effort should be dedicated to anchor-free and rehearsal-free object detection. In this paper, we propose MultIOD, a class-incremental object detector based on CenterNet. Our  contributions are: (1) we propose a multihead feature pyramid and multihead detection architecture to efficiently separate class representations, (2) we employ transfer learning between classes learned initially and those learned incrementally to tackle catastrophic forgetting, and (3) we use a class-wise non-max-suppression as a post-processing technique to remove redundant boxes. Results show that our method outperforms  state-of-the-art methods on two Pascal VOC datasets, while only saving the model in its current state, contrary to other distillation-based counterparts.
\end{abstract}    
\section{Introduction}
\label{sec:introduction}


Catastrophic forgetting \cite{mccloskey1989_catastrophic} is a significant challenge when artificial agents update their knowledge with new data. It involves losing past knowledge while rapidly transforming the model representations to fit the new data distribution.  In scenarios where data arrives in streams, requiring ongoing model adaptation, the concept of Continual Learning (CL); i.e., the ability to learn from new examples while not forgetting knowledge from the old ones, gains prominence. Class-incremental learning is a subdomain of Continual Learning, where new classes are added at each system update (called state). It gained an increasing interest in the last few years due to the emergence of different deep learning algorithms \cite{rebuffi2017_icarl, shmelkov2017_ilod, hayes2020_rodeo, dongbao2022_rd_iod, masana2022_study, liu2021_incdet}. Class-Incremental Object Detection (CIOD) is  interesting in practice. It can be deployed in autonomous cars that continuously explore objects on road \cite{shieh2020_continual}, or in security cameras to easily detect infractions, or even in large events to continuously capture attendees density for statistical purposes \cite{chaudhari2018_density}. In computer vision, class-incremental learning is usually applied to classification, object detection, and segmentation. Rehearsal-free continual object detection and segmentation comprise an additional challenge compared to classification. The absence of annotations for objects belonging to earlier classes leads to their classification as background  \cite{angelo2022_review}. This phenomenon is called \textit{background interference} (or shift) and it  aggravates the effect of catastrophic forgetting.

\begin{figure}[t]
	\begin{center}
    \includegraphics[width=0.4\textwidth]{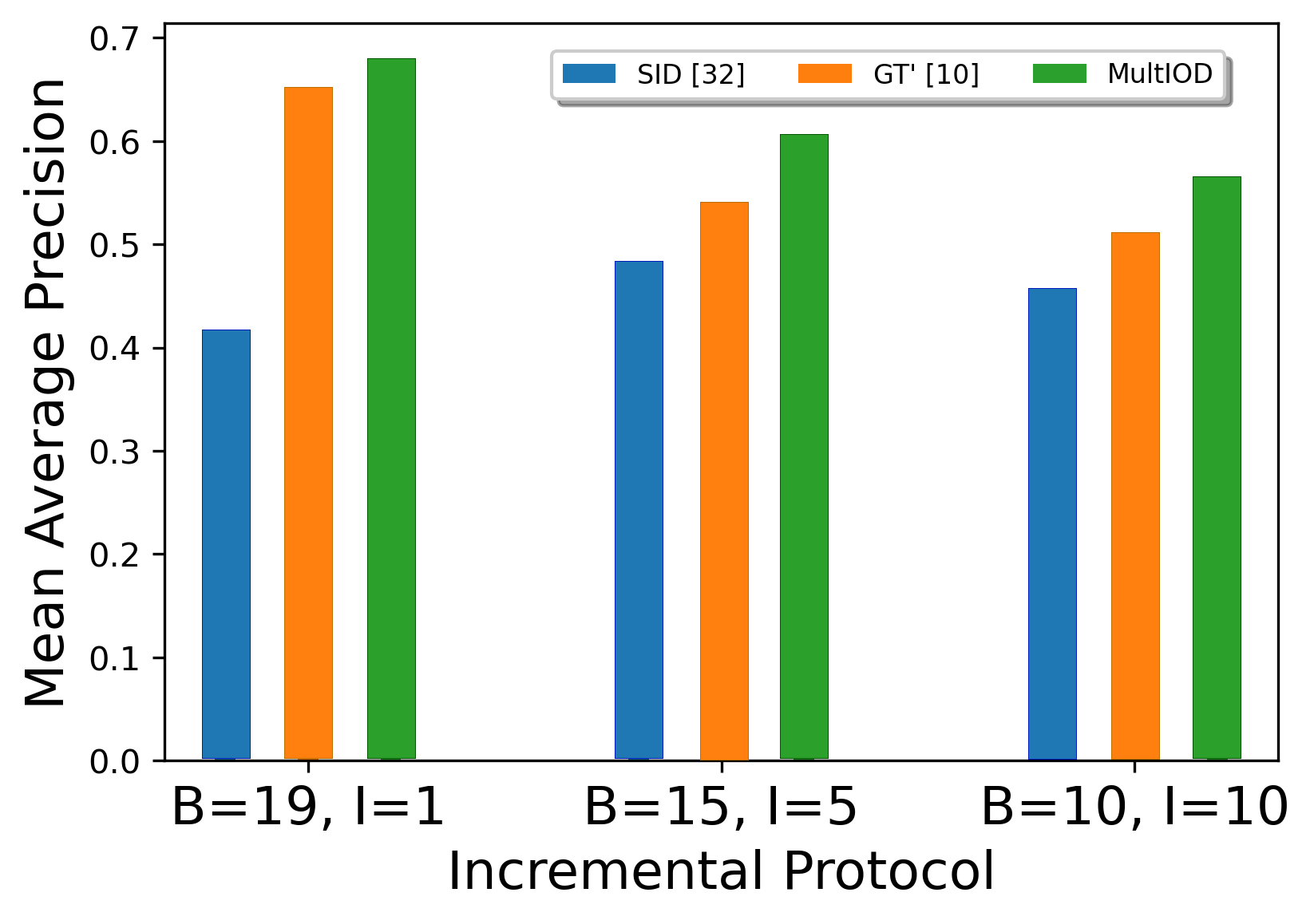}
    \vspace{-0.5cm}
	\end{center}
    \vspace{-1em}
	\caption{Mean Average Precision (IoU=0.5) on VOC0712  using different number of base classes ($B$) and incremental classes ($I$).}
     \vspace{-1em}
	\label{fig:main_results}
\end{figure}


CIOD community proposed some methods to tackle the problem \cite{angelo2022_review}. However, most of them are not suitable for scenarios where real-time adaptation is required. Indeed, on the one hand, most existing CIOD  models \cite{shmelkov2017_ilod, chen2019_new, liu2021_incdet, peng2020_faster_ilod, ramakrishnan2020_RKT} are based on Faster-RCNN \cite{ren2015_faster_rcnn}, a two-stage detection algorithm. 
Although this method has been proven useful for object detection, it comes at the expense of running speed. 
On the other hand, the largest room was booked for rehearsal-based methods, where past data is rehearsed to refresh the model representation and tackle forgetting. However, this scenario is inapplicable in cases where access to past data is impossible due to privacy issues or  hardware limitations.

In this paper, we push the effort towards developing continual object detectors that are anchor-free and rehearsal-free. This is indeed the most challenging scenario, but the most useful in practice. Therefore, we propose \textit{Multihead Incremental Object Detector} ($MultIOD$), a new CIOD model based on CenterNet algorithm \cite{zhou2019_centernet}. These are our main contributions:

1.  \underline{Architecture:} we propose a multihead feature pyramid \cite{lin2016_fpn} to share upsampling layers of a group of classes that appear in the same state, while a multihead detector is used to efficiently separate class representations (Subsec \ref{subsec:multihead_centernet}).

2.  \underline{Training:} we apply a transfer learning between classes learned in the initial state and classes learned incrementally to tackle catastrophic forgetting (Subsec \ref{subsec:transfer_learning}).

3. \underline{Inference:} we use a class-wise non-max-suppression as a post-processing technique to remove redundant boxes within the same class (Subsec \ref{subsec:inference_process}).

Figure \ref{fig:main_results} shows
that our method outperforms SID \cite{peng2020_sid} and GT' \cite{gang2022_pdm}, two distillation-based methods that are built on top of CenterNet. Note that our method only requires saving the model at the current state, while other state-of-the-art distillation-based counterparts need both the current and previous models to learn. Also,  our method trains faster since the number of trainable parameters at each state is reduced thanks to the transfer learning scheme.



\begin{figure*}[t]
	\begin{center}
\includegraphics[width=\textwidth]{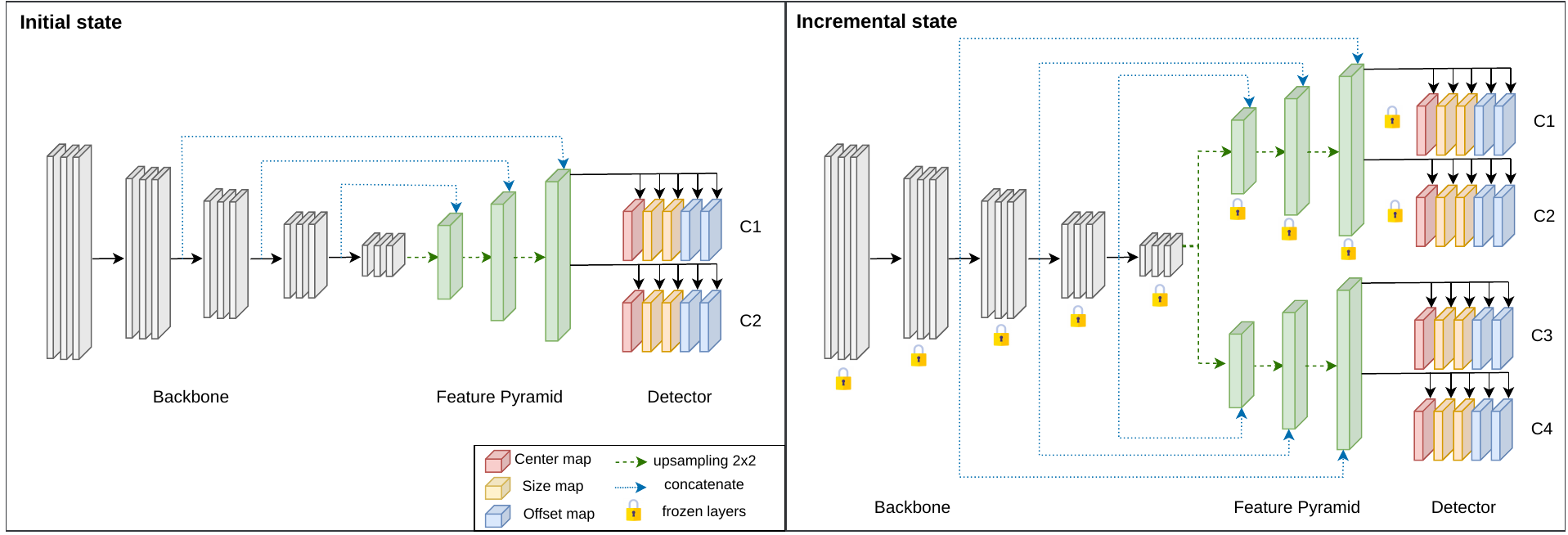}
	\end{center}

\vspace{-2em}
 
	\caption{Illustration of $MultIOD$, depicting two toy states: one initial state (on the left), and one incremental state (on the right). The model is trained classically in the initial state using data from classes C1 and C2, while in the incremental state, it is updated using data from classes C3 and C4 only. The backbone, the feature pyramid of classes C1 and C2, as well as their detection heads are frozen once these classes are learned. Only the feature pyramid of classes C3 and C4, and their detection heads are trained in the incremental state.}
 \vspace{-1em}
	\label{fig:overview}
\end{figure*}

\section{Related Work}
\label{sec:related_work}

Continual Learning is a hot topic of research, but most existing methods tackle classification task, and very few effort is dedicated to object detection and semantic segmentation. In this paper, we push the effort towards providing more solutions for continual object detection. We categorize existing methods into three groups as in \cite{masana2022_study}. 

\subsection{Fine-Tuning-based approaches}
Here, model parameters are continuously updated at each incremental state.  ILOD \cite{shmelkov2017_ilod} is one of the first works that tackled CIOD problem. Authors propose a loss that balances the interplay between predictions of new classes, while they use a distillation loss to minimize the discrepancy between responses for past classes from the previous and the current models.
\cite{chen2022_remote} and \cite{hao2019_end_to_end} both modify the Region Proposal Network (RPN) of a Faster-RCNN \cite{ren2015_faster_rcnn} to accommodate new classes. The former uses a fully connected layer along with distillation to classify proposals. The latter adds a knowledge distillation loss from a teacher model using a domain-specific dataset. Similarly to \cite{chen2022_remote}, the authors of \cite{zhou2020_lifelong} use distillation  with a sampling strategy that helps better select proposals of foreground classes. \cite{ramakrishnan2020_RKT} distills knowledge for proposals that have a strong relation with ground-truth boxes. \cite{ulhaq2022_yolov3_smart} also uses distillation of output logits, but with a YoloV3 backbone. Other works distill intermediate features instead of logits. \cite{chen2019_new} proposes a Hint Loss to tackle catastrophic forgetting, and uses an adaptive distillation approach that uses both RPN output and features. 

Rehearsal of past class exemplars is widely used to tackle catastrophic forgetting. It was initially proposed in \cite{rebuffi2017_icarl} to classify images incrementally. Its effectiveness was proven in classification on multiple datasets \cite{belouadah2020_study, masana2022_study}. In object detection, it was used by \cite{hao2019_take_goods} and \cite{shieh2020_continual} who combined it with rehearsal and distillation, respectively. Meta-ILOD \cite{joseph2020_meta_ilod} uses both distillation and rehearsal to tackle forgetting. In addition, it uses a gradient-conditioning loss for the region of interest component as a regularization strategy. In \cite{liu2020_multi}, authors propose an attention mechanism for feature distillation and an adaptive exemplar selection strategy for rehearsal. \cite{zhang2020_dmc} uses an external dataset and distills knowledge from two networks that learn past and new classes, respectively, to a new separate network. Alternatively, RILOD \cite{li2019_rilod} helps in collecting and annotating data from search engines, and using it in incremental detection. Finally, IncDet \cite{liu2021_incdet} builds on Elastic Weights Consolidation (EWC) and uses a pseudo-labeling technique in addition to a Huber loss for regularization. The pseudo labeling is widely used by the community to tackle background interference. We provide in Appendix \textcolor{red}{8} a figure that explains it.

There are few CenterNet-based CIOD models \cite{gang2022_pdm, peng2020_sid}. GT' \cite{gang2022_pdm} combines the ground-truth bounding boxes with predictions of the model in the previous state. The latter are first converted to bounding boxes. Then, redundant boxes are cleaned before to be distilled to the current model. Selective and Inter-related Distillation (SID) \cite{peng2020_sid} is another continual detector where the authors propose a new distillation method. It consists in using Euclidean distance to measure the inter-relations between instances. This helps in considering the relationship among different instances, and thus improves the transfer. We compare $MultIOD$ to both methods. Note that SID and GT' save both previous and current models in order to perform distillation, while  $MultIOD$ needs the current model only. 

\subsection{Parameter-Isolation-based approaches}
This type of approaches uses a subset of the model parameters to learn a different set of classes each time.

MMN \cite{li2018_mmn} freezes the most important weights in the neural network and uses the least useful ones to learn new classes. Weights importance is computed based on their magnitudes that are evaluated after each incremental state. Alternatively, \cite{zhang2021_merge_expert} uses pruning techniques \cite{liu2017_pruning} to remove useless channels and residual blocks, while it uses a YoloV3-based ensemble network to perform detection.

$MultIOD$ incorporates features of a parameter-isolation technique. We assign a dedicated detection head to each class while sharing a common feature pyramid among classes within the same state. The backbone remains the only component shared across the detector.

\vspace{-0.5em}

\subsection{Fixed-Representation-based approaches}


These approaches are based on a frozen feature extractor to transfer knowledge between the initial and incremental classes.  In RODEO \cite{hayes2020_rodeo}, authors freeze the neural network after learning the first batch of classes. Later, they quantize the extracted features in order to have compact image representations. The latter are rehearsed in incremental states. 

    
Transfer learning was proven to cope well with class-incremental learning when no memory of the past is allowed. \cite{belouadah2018_deesil} trains a deep feature extractor on initial classes and a set of Support Vector Machines (SVMs) \cite{boser1992_svms} is used to learn classes incrementally. $MultIOD$ shares the same spirit as \cite{belouadah2018_deesil}, where the CenterNet backbone is shared between all classes, and a different detection head is specialized to learn each class. $MultIOD$ is also considered as a fixed-representation-based algorithm. 


Most of the works from the literature are built on top of a Faster-RCNN object detector. However, this detector cannot run in real-time. Real-life applications require a faster model to learn continuously while acquiring data in streams. According to a recent survey \cite{arani2022_survey}, keypoint-based detectors are the fastest, while being efficient. That is why we choose CenterNet \cite{zhou2019_centernet} as a main algorithm to develop our class-incremental detector $MultIOD$. 
While multihead CenterNet has been investigated previously in \cite{heuer2021_centernet_multitask}, its primary focus was on accommodating multi-task objectives such as object detection, pose estimation, and semantic segmentation, instead of continual learning.


\section{Proposed Method}
\label{sec:proposed_method}
\vspace{-0.5em}
\subsection{Problem Definition}
\label{subsec:problem_definition}

Let's consider a set of states $\mathcal{S}= \{S_0, S_1, ..., S_{n-1}\}$, where $n$ is the number of states. The initial state $S_0$ contains $B$ classes, and incremental states $S_{i>0}$ contain $I$ classes each. In a general form, we note $|\mathcal{C}_i|$ the total number of classes seen so far in a state $S_i$. $\mathcal{D}= \{D_0, D_1, ..., D_{n-1}\}$ are the sets of images of each state (these sets can overlap).  An initial model $M_0$ is trained classically on $D_0$ which contains the first $B$ classes. Incremental models $M_1$, $M_2$, ..., $M_{n-1}$ are trained on $D_1$, $D_2$, ..., $D_{n-1}$  in states $S_1$, $S_2$, ..., $S_{n-1}$, respectively. Note that objects from $S_i$ can appear in $D_{j>i}$ and objects from $D_{j>i}$ can be present in $S_{i}$ but, each time, only objects from classes of the corresponding state $S_i$ are annotated (objects from other states are not). This is known as background interference (explained in Appendix \textcolor{red}{8}), a phenomenon which augments the complexity of class-incremental learning for object detection.

At each incremental state $S_{i>0}$, the model $M_i$ is initialized with weights of model $M_{i-1}$. $M_i$ has access to all training ground-truth bounding boxes from new classes, but no bounding box from past classes is available. When testing, annotations from all classes $|\mathcal{C}_i|$ are available to evaluate the performance of the model on all data seen so far. Having no access to past data during training is the most challenging scenario in class-incremental learning, yet the most interesting one in practice. 

An object detection model usually comprises:

 $\bullet$ \textbf{A Backbone:} a classification network without the final dense layer. \underline{Example:} ResNet \cite{he2015_resnet}, EfficientNet \cite{tan2019_effnet}, etc.

\hspace{-0.2cm} $\bullet$ \textbf{An Upsampling Network:} receives the output from the backbone and increases its dimensions to generate final prediction maps with enhanced resolution. \underline{Example:} Deformable Convolutions \cite{dai2017_dcn}, Feature Pyramids \cite{lin2016_fpn}, etc.
    
\hspace{-0.2cm} $\bullet$ \textbf{A Detection Network:} takes the output of the upsampling network and makes the final prediction. \underline{Example:} CenterNet \cite{zhou2019_centernet}, Yolo \cite{redmon2016_yolo}, SSD \cite{wei2015_ssd}, etc.

Since $MultIOD$ is based on CenterNet, we briefly remind its main components in the next subsection.

\subsection{CenterNet Object Detector}
CenterNet \cite{zhou2019_centernet} is an anchor-free object detector. It considers objects as points, and it outputs three types of maps: 

\vspace{-1em}

\paragraph*{ - Center map:} encodes the center of the objects using a Gaussian distribution, where the mean corresponds to the object center location, the peak value is set to 1, and the standard deviation varies based on the size of the object. The focal loss is used in \cite{zhou2019_centernet} as an objective (Equation \ref{eq:center}):

\begin{equation}
\mathcal{L}_{focal} = \frac{-1}{N} \sum_{xyc} \left\{ \begin{array}{rl}
(1-\Hat{Y}_{xyc})^\alpha ~~ log(\Hat{Y}_{xyc}) & \mbox{if}~ Y_{xyc}=1\\

(1-Y_{xyc})^\beta ~~ (\Hat{Y}_{xyc})^\alpha  & \mbox{otherwise}\\
 log(1-\Hat{Y}_{xyc}) & 

\end{array}\right.
\label{eq:center}
\end{equation}

where $x$ and $y$ are center map coordinates, $c$ is the class index, $\alpha, \beta$ are parameters of the focal loss. $N$ is the number of objects in the image. $\Hat{Y}$ is the predicted center map, and $Y$ is the ground-truth center map.

\paragraph*{ - Size map:} consists of two maps (for width and height). At the location corresponding to the peak of each Gaussian (center of the object), the width of this object is inserted in the width map, and its height in the height map. The size loss in \cite{zhou2019_centernet} is an L1 loss (Equation \ref{eq:size}):

\begin{equation}
\mathcal{L}_{size} = \frac{1}{N} \sum_{i=0}^{N-1} |\Hat{S_i} - S_i|
\label{eq:size}
\end{equation}

where  $\Hat{S}$ is the predicted size map, and $S$ is the ground-truth size map. Note that the loss is only computed on map coordinates corresponding to the center of the objects.

\vspace{-1em}

\paragraph*{ - Offset map:} consists also of two maps (for x and y axes). It is used to recover the discretization error caused by  output stride when encoding object centers. \cite{zhou2019_centernet} use an L1 loss, and they compute it on centers only (Equation \ref{eq:offset}):

\begin{equation}
\mathcal{L}_{offset} = \frac{1}{N} \sum_{i=0}^{N-1} |\Hat{O_i} - O_i|
\label{eq:offset}
\end{equation}

where  $\Hat{O}$ is the predicted offset map, and $O$ is the ground-truth offset map. The overall loss is the combination of the three losses (Equation \ref{eq:centernet}):

\begin{equation}
\mathcal{L} = \lambda_{focal} \times \mathcal{L}_{focal}  +  \lambda_{size} \times \mathcal{L}_{size} +  \lambda_{offset} \times \mathcal{L}_{offset}
\label{eq:centernet}
\end{equation}

where $\lambda_{focal}=1.0$, $\lambda_{size}=0.1$, $\lambda_{offset}=1.0$ \cite{zhou2019_centernet}. 

Note that CenterNet uses one center map per class, and shared size and offset maps between all classes.


\subsection{Multihead CenterNet for Class Separability}
\label{subsec:multihead_centernet}

We build on CenterNet to develop our class-incremental detector. The central concept driving $MultIOD$ is to differentiate the parameters responsible for learning distinct classes. This differentiation is not about complete separation, but rather about determining which parameters should be shared and which should be selectively distinguished. Therefore, we propose the creation of a multihead architecture, which we will elaborate on hereafter:


\vspace{-0.75em}

\paragraph*{- Multihead Feature Pyramid:} 
Feature Pyramid Network (FPN) \cite{lin2016_fpn} helps to build high-level semantic feature maps at all scales. This network is efficient for detecting both large and small objects. In an incremental scenario, we propose to use one feature pyramid for each group of classes that occur at the same time in the data stream. This helps in sharing a subset of parameters between these classes and reinforces their learning. The detailed architecture of our FPN is in Appendix \textcolor{red}{10}. All implementation details are in Appendix \textcolor{red}{9}.

\vspace{-0.75em}

\paragraph*{- Multihead Detector:} 
We adapt the original CenterNet \cite{zhou2019_centernet} architecture to a class-incremental scenario as follows: (1) In contrast to \cite{zhou2019_centernet} that shares size and offset maps between classes, we use one size map and one offset map for each class. The motivation behind this choice is to enable the separation of class representations in the architecture. Furthermore, the original CenterNet, by definition \cite{zhou2019_centernet}, lacks the capability to address scenarios in which two objects of distinct classes happen to share the exact same center position. This limitation results in the necessity to encode either one object or the other in the size map, but not both simultaneously. 


\vspace{-0.5em}

\subsection{Transfer Learning to Tackle Forgetting}  
\label{subsec:transfer_learning}

In a rehearsal-free protocol, updating our model using only new data will result in a significant bias towards these specific classes in the weights of the backbone, feature pyramids, and detection heads. This bias leads to catastrophic forgetting, even when a multihead architecture is employed. To mitigate this, we propose a strategy that consists in freezing the backbone, feature pyramids of previously learned classes, and their corresponding detection heads once they are learned. This strategy efficiently minimizes the distribution shift of model parameters, facilitating effective transfer between the classes learned during the initial state and those learned incrementally. Transfer learning was proven to be effective on rehearsal-free continual learning for both classification \cite{belouadah2018_deesil} and detection \cite{hayes2020_rodeo}. It is noteworthy to mention that freezing a large part of the neural network leads to faster training since we have fewer parameters to optimize.




Figure \ref{fig:overview} provides an overview of our method. It depicts two states: one initial, and one incremental. In the initial state, since we have only one group of classes, we use one feature pyramid shared between these classes, and one detection head per class. We perform a classical training with all data  from these classes. In the incremental state, we freeze the backbone, the feature pyramid of past classes and their detection heads. Then, we add a new feature pyramid and detection heads to learn the new classes.

\paragraph*{Training Objective}

Since we use one separate size map and offset map for each class, we modify their respective losses, as in Equations \ref{eq:size_multihead} and \ref{eq:offset_multihead}. 

\begin{equation}
\mathcal{L}_{size}' = \frac{1}{N} \sum_{c=0}^{|\mathcal{C}|-1}  \sum_{i=0}^{N_c-1} |\Hat{S}_{ic} - S_{ic}|
\label{eq:size_multihead}
\end{equation}

where $|\mathcal{C}|$ is the number of classes seen so far, and $N = N_0 + N_1 +...+N_{|\mathcal{C}|-1}$ is the total number of objects in the image. Similarly for the offset loss:

\begin{equation}
\mathcal{L}_{offset}' = \frac{1}{N} \sum_{c=0}^{|\mathcal{C}|-1}  \sum_{i=0}^{N_c-1} |\Hat{O}_{ic} - O_{ic}|
\label{eq:offset_multihead}
\end{equation}



\subsection{Class-wise Non-Max-Suppression}
\label{subsec:inference_process}

As mentioned in Section \ref{subsec:problem_definition}, the model is evaluated on all classes seen so far. In object detection, the bounding boxes generation is usually followed by a Non-Max-Suppression (NMS) operation, in which redundant boxes are eliminated, and only the most pertinent ones are kept. However, in CenterNet \cite{zhou2019_centernet}, authors demonstrate that NMS is not useful in their algorithm. The latter directly predicts object centers, resulting in a situation where each object corresponds to just one positive anchor in the ground truth.

In $MultIOD$, however, we found it very useful to use NMS and we will provide empirical evidence in Table \ref{tab:ablation_nms_mnist} to support our finding. We select to use class-wise NMS in our model for two reasons. First, since the backbone is frozen after learning the first set of classes, the model tends to favor predictions of past classes whose confidence scores are higher than those of new classes. Second, it is important to not remove boxes belonging to different classes but share the same location. Algorithm \ref{algorithm:nms} details class-wise NMS.


\begin{algorithm}
  \caption{Class-wise Non-Maximum Suppression}
  \label{algorithm:nms}
  \begin{algorithmic}[1]
    \Function{ClassWiseNMS}{\textit{detections}, \textit{threshold}}
      \State Sort \textit{detections} by decreasing confidence scores \\~~~~~~~~~~within each class
      \State Initialize an empty list \textit{selected-dets}
      \For{\textit{class} \textbf{in} \textit{classes}}
        \State \textit{class-dets} $\gets$ detections of class
        \While{\textit{class-dets} is not empty}
          \State \textit{max-det} $\gets$ detection with highest\\~~~~~~~~~~~~~~~~~~~~~~~~~~~~~~~~~~~~ confidence in class-dets
          \State Add \textit{max-det} to \textit{selected-dets}
          \State Remove \textit{max-det} from \textit{class-dets}
          \For{\textit{det} \textbf{in} \textit{class-dets}}
            \If{IoU(\textit{det}, \textit{max-det}) $\geq$ \textit{threshold}}
              \State Remove \textit{det} from \textit{class-dets}
            \EndIf
          \EndFor
        \EndWhile
      \EndFor
      \State \textbf{return} \textit{selected-dets}
    \EndFunction
  \end{algorithmic}
\end{algorithm}

\section{Experiments}
\label{sec:experiments}

\subsection{Compared Methods}
\label{subsec:baselines}

We compare $MultIOD$ with CenterNet-based methods:

\hspace{-0.2cm} $\bullet$ \textbf{Vanilla Fine-tuning (FT)} - the lower-bound method in which we directly fine-tune CenterNet without any specific strategy to tackle catastrophic forgetting.

\hspace{-0.2cm} $\bullet$ \textbf{Learning without Forgetting (LwF)} \cite{li2016_lwf} - the previous model is distilled to guide the training of the current model. Note that this method was initially proposed for classification.

\hspace{-0.2cm} $\bullet$ \textbf{SID} \cite{peng2020_sid} - this is a distillation-based method, where authors distill intermediate features between the past and current model, and also distances between samples. The method was applied on both CenterNet and FCOS \cite{tian2019_fcos} detectors, but only the former is used here for comparability.

\hspace{-0.2cm} $\bullet$ \textbf{SDR} \cite{michieli2021_sdr} - this method was initially proposed for semantic segmentation. It is based on three main components: prototype matching, feature sparsification, and contrastive learning. We try two versions of this method: SDR1 distills only the center map, while SDR2 distills all maps.
    
\hspace{-0.2cm} $\bullet$ \textbf{GT'} \cite{gang2022_pdm} - instead of directly distilling CenterNet maps from the previous to the current model, the past model is used to annotate data from new classes. After extracting the bounding boxes, the latter are encoded along with the ground truth before training the model on both the ground-truth and the pseudo-labeled data. Please note that the experimental protocol of this method is not realistic because authors remove images where past class objects are present.
    
\hspace{-0.2cm} $\bullet$ \textbf{Full} - the upper-bound method. A full training is performed on all classes with all their available data.

Note that we take results of $LwF$ and $SDR$ from \cite{peng2020_sid} and \cite{michieli2021_sdr}, respectively.

For a fair comparison, we only compare our method with algorithms that do not perform any rehearsal, but we still report results against some two-stage continual detectors that use or not rehearsal, in Appendix \textcolor{red}{13}. While all the aforementioned methods build upon CenterNet with a ResNet50 backbone, we opt to use the EfficientNet family of models as backbone because they are more suitable for efficient training. Unless specified otherwise, our choice of backbone is EfficientNet-B3 because it provides the best trade-off between performance and number of parameters. CenterNet with ResNet50 contains about 32M parameters, while CenterNet with EfficientNet-B3 model contains 17.7M parameters only. That is, we reduce around 45\% of the number of parameters. We will prove in the results section, that even though our model achieves lower results in classical training (with all data available at once), it still outperforms by a good margin the other methods in almost all configurations. Our model is not only more robust against  forgetting, but it is also parameter-efficient.

\subsection{Datasets}
\label{sub:datasets}


    

\hspace{0.2cm} $\bullet$ \textbf{MNIST-Detection} (MNIST below) - this dataset was initially designed for digit classification \cite{lecun2010mnist}. We create an object-detection toy version of it to perform our ablation studies as it runs faster than other datasets.  We used this github repository \footnote{https://github.com/hukkelas/MNIST-ObjectDetection} to create training and validation sets of 5000 and 1000 images of size 512$\times$512, respectively. We made sure to create a challenging dataset. The dataset creation procedure is detailed in Appendix \textcolor{red}{11}.

\hspace{-0.2cm} $\bullet$ \textbf{Pascal VOC2007} \cite{everingham2010_pascal_voc} - this dataset contains 20 object classes with 5,011 and 4,952 training and validation examples, respectively. Note that we use both the training and validation sets for training, as in  \cite{peng2020_sid}. 

\hspace{-0.2cm} $\bullet$ \textbf{Pascal VOC0712} \cite{everingham2010_pascal_voc}  - here  we use both the training and validation sets of VOC 2007 and 2012 as in  \cite{gang2022_pdm, peng2020_sid, zhou2019_centernet}. We use the test set of VOC2007 for validation. In total, this dataset contains 16,550 and 4,952 training and validation images, respectively.

\subsection{Methodology}
\label{subsec:methodology}
We first apply most of the ablations on MNIST dataset, then perform the final experiments on VOC2007 and VOC0712.
Following \cite{peng2020_sid, angelo2022_review, joseph2020_meta_ilod},
We evaluate our method using three incremental learning scenarios: we order classes (numerically for MNIST, alphabetically for VOC), then divide them into two states. For MNIST, we use $B=9, I=1$; $B=7, I=3$ and $B=5, I=5$. We use the following number of classes per state for the VOC dataset: $B=19, I=1$; $B=15, I=5$ and $B=10, I=10$. Varying the number of classes between states is important to assess the robustness of CL methods.

\subsection{Evaluation Metrics}
\label{subsec:metrics}


\hspace{0.2cm} $\bullet$ \textbf{mAP@0.5} - the mean average precision computed at an IoU (Intersection-over-Union) threshold of 0.5. 
    
\hspace{-0.2cm} $\bullet$ \textbf{mAP@[0.5, 0.95]} - the mean average precision averaged on IoU threshold that varies between 0.5 and 0.95 with a step of 0.05. Note that results with this metric are presented in Appendix \textcolor{red}{12}.

\hspace{-0.2cm} $\bullet$ $\mathbf{F_{mAP}}$ - the harmonic mean between the mean average precision of past and new classes.

    \begin{equation}
        F_{mAP} = 2 \times \frac{mAP_{past} \times mAP_{new}}{mAP_{past} + mAP_{new}}
    \end{equation}

When $F_{mAP}=0.0$, this means that the model completely failed to detect either past classes or new classes. 


\section{Ablation Studies} 
\label{subsec:ablation}

\subsection{Ablation of Backbones}
The backbone has a direct impact on model performance because it is responsible for the quality of extracted features. In Table \ref{tab:ablation_backbones}, we vary the backbone and use EfficientNet-B0 (10 million params), EfficientNet-B3 (17.7 million params), and EfficientNet-B5 (37 million params) on VOC2007 and VOC0712. We provide results on MNIST in Appendix \textcolor{red}{13}. Results show that the best overall backbone is EfficientNet-B3. EfficientNet-B0 is smaller and has more difficulties to generalize. However, it still provides decent performances while having 10M params only. EfficientNet-B5 provides comparable results with EfficientNet-B3. We decided to keep the latter since it has the best compromise between parameters number and performance.

\begin{table*}[htb!]
	\begin{center}
		\resizebox{\textwidth}{!}{
			\begin{tabular}{|l|c|c|c|c|c|c|c|c|c|c|c|c|c|c|c|}
				\hline
    & \multicolumn{7}{c|}{VOC2007} & \multicolumn{7}{c|}{VOC0712} \\ \hline
    Method & Full  & \multicolumn{2}{c|}{$B=19$, $I=1$} & \multicolumn{2}{c|}{$B=15$, $I=5$} & \multicolumn{2}{c|}{$B=10$, $I=10$} & Full & \multicolumn{2}{c|}{$B=19$, $I=1$} & \multicolumn{2}{c|}{$B=15$, $I=5$} & \multicolumn{2}{c|}{$B=10$, $I=10$} \\	
				\hline
    
    & $mAP$  & $mAP$ & $F_{mAP}$ & $mAP$ & $F_{mAP}$ & $mAP$ & $F_{mAP}$ & $mAP$ & $mAP$ & $F_{mAP}$ & $mAP$ & $F_{mAP}$ & $mAP$ & $F_{mAP}$\\ \hline
    EfficientNet-B0  &  56.7  & 55.7 &  35.6 & 49.2 & 33.8 & 46.3 & 45.5  &  66.9 & 65.0 & 54.0 & 57.7 &  44.7 & 53.9 &  52.9 \\
    EfficientNet-B3  &  60.4   & \textbf{59.9} & 39.2 & \textbf{52.6} & \textbf{37.8} & \textbf{48.4} & \textbf{47.2} &  69.5 & 68.0 & \textbf{56.9} & 60.7 &  \textbf{47.0} & \textbf{56.6} &  \textbf{55.8} \\ 
    EfficientNet-B5 &   \textbf{61.3}   & \textbf{59.9} &  \textbf{45.2} & 52.5 & 36.5 & 45.2 &  42.4 &  \textbf{70.3} & \textbf{68.4} & 54.7 & \textbf{60.8} &  46.5 & 55.3 &  53.9 \\   \hline     
			\end{tabular}
		}
	\end{center}
	\vspace{-1.5em}
	\caption{Ablation of backbones on VOC2007 and VOC0712 datasets (mAP@0.5).}
    \vspace{-1em}
	\label{tab:ablation_backbones}
\end{table*}


\subsection{Ablation of Frozen Layers}
One of the main components of our method is to use transfer learning between classes learned in the initial state, and classes that are learned incrementally. Therefore, studying the impact of freezing the different layers of our system is a crucial part to assess its robustness. In Table \ref{tab:ablation_freeze}, we use MNIST dataset and EfficientNet-B0 backbone in a setting where $B=5$ and $I=5$, and test the following configurations: (1) do not freeze anything in the architecture, (2) freeze the backbone only, (3) freeze both the backbone and the feature pyramid of past classes, and (4) our full method, in which we freeze not only the backbone, but also the feature pyramid of past classes, and their detection heads. Results show that it is crucial to freeze past class detection heads in order to avoid their catastrophic forgetting. In fact, it is meaningful to freeze them only if all the preceding components are also frozen (the feature pyramid of past classes and the backbone).

\begin{table}[H]
	\begin{center}
		\resizebox{0.42\textwidth}{!}{
			\begin{tabular}{|l|c|c|c|c|c|}
				\hline
	backbone & feature pyramid & detection head & $mAP$ & $F_{mAP}$ \\	
				\hline
    $\times$ &  $\times$ &  $\times$ & 44.2 & 0.0  \\		
    \faLock &  $\times$ & $\times$ & 44.3 & 0.0  \\		
    \faLock & \faLock &  $\times$ & 42.0 & 0.0  \\
    \faLock & \faLock & \faLock & \textbf{91.3} & \textbf{91.3}\\	
    \hline                
    \end{tabular}
		}
	\end{center}
	\vspace{-1.5em}
	\caption{Performance of $MultIOD$ on MNIST with EfficientNet-B0 and $B=5, I=5$ when ablating its components freezing.}
 \vspace{-1em}
	\label{tab:ablation_freeze}
\end{table}

\subsection{Ablation of Non-Max-Suppression Strategies}
Using multiple detection heads leads to the risk of predicting multiple objects (from different classes) at the same pixel location. Thus, it is important to eliminate irrelevant bounding boxes (false positives). In this experiment, we ablate the following NMS strategies:

\hspace{-0.2cm} $\bullet$ No-NMS: this is the method originally used in CenterNet \cite{zhou2019_centernet}, where no non-max-suppression is applied.

\hspace{-0.2cm} $\bullet$ Inter-class NMS: here, a standard NMS algorithm is utilized to eliminate redundant bounding boxes, irrespective of their class membership.

\hspace{-0.2cm} $\bullet$ Class-wise NMS: as described in Subsection \ref{subsec:inference_process}, it is designed to enhance the precision of boxes selection within individual classes, effectively curbing redundancy and selecting the most pertinent boxes for each specific class.

\hspace{-0.2cm} $\bullet$ Soft-NMS: unlike traditional NMS, which remove boxes that overlap significantly, it applies a decay function to the confidence scores of neighboring boxes, gradually reducing their impact. This results in a smoother suppression of redundant boxes and helps retain boxes with slightly lower scores that might still contribute to accurate detection.

Table \ref{tab:ablation_nms_mnist} shows that the method which provides the best results is class-wise NMS, the one we choose to use. The second best method is the standard inter-class NMS. The former helps in detecting more objects, specifically when objects of different classes are highly overlapped. This comes at the expense of having some false positives. In contrast, inter-class NMS reduces the number of false positives at the expense of not detecting bounding boxes of different classes that are at the same location in the image. Depending on the use case, one or the other method can be preferred. In our use case, Soft-NMS did not provide good results. This can be explained by the fact that the gradual decay of confidence scores in Soft-NMS might inadvertently allow boxes with lower scores to persist, potentially leading to false positives or less accurate detections. Finally, unsurprisingly, the method with the worst performance is No-NMS due to the high number of false positives. Note that the same ablation for VOC2007  is in Appendix \textcolor{red}{14}.

\begin{table}[htb!]
	\begin{center}
		\resizebox{0.5\textwidth}{!}{
			\begin{tabular}{|l|c|c|c|c|c|c|c|c|}
				\hline
    Method & Full  & \multicolumn{2}{c|}{$B=9$, $I=1$} & \multicolumn{2}{c|}{$B=7$, $I=3$} & \multicolumn{2}{c|}{$B=5$, $I=5$}  \\	
				\hline
    
    & $mAP$  & $mAP$ & $F_{mAP}$ & $mAP$ & $F_{mAP}$ & $mAP$ & $F_{mAP}$ \\ \hline
    No-NMS &  90.2 & 89.5 & 89.7 &  90.8 &  91.3 & 88.4 &  88.2 \\ 
    Soft-NMS  &   91.5 & 89.7 & 89.5 &  91.1 &  91.4 & 88.5 &  88.4  \\ 
    Inter-class NMS &   91.8 & 90.2 & 90.2 & 91.7 & 92.0 &  89.9 &  89.8 \\
    Class-wise NMS &   \textbf{93.1} & \textbf{91.3} & \textbf{91.2} &  \textbf{93.1} &  \textbf{93.5} & \textbf{91.3} &  \textbf{91.3}  \\ \hline   
     
			\end{tabular}
		}
	\end{center}
	\vspace{-1.5em}
	\caption{Performance of our model using MNIST  dataset with different NMS strategies and EfficientNet-B0.}
	\vspace{-1em}
 \label{tab:ablation_nms_mnist}
\end{table}

\section{Results and Discussion}
\label{sec:results}

Tables \ref{tab:results_voc2007} and \ref{tab:results_voc0712} provide the main results of $MultIOD$ compared to state-of-the-art methods on VOC2007 and VOC0712 datasets, respectively. 

\begin{table}[htb!]
\setlength{\tabcolsep}{3pt}
	\begin{center}
		\resizebox{0.5\textwidth}{!}{
			\begin{tabular}{|l|c|c|c|c|c|c|c|c|}
				\hline
	Method & Full & \multicolumn{2}{c|}{$B=19$, $I=1$} & \multicolumn{2}{c|}{$B=15$, $I=5$} & \multicolumn{2}{c|}{$B=10$, $I=10$} \\	
				\hline
     & $mAP$   & $mAP$ & $F_{mAP}$ & $mAP$ & $F_{mAP}$ & $mAP$ & $F_{mAP}$ \\ \hline
   	FT &  \multirow{3}{*}{65.6}  & 7.3 &  9.0 & 13.2 & 5.0 & 28.6 &  0.0   \\		
    LwF \cite{li2016_lwf} &  & 12.5 & 10.9 & 6.9 & 1.5 & 17.6 & 0.0 \\	
    SID \cite{peng2020_sid} &   & 45.5 & 27.0  & 51.9 & \textbf{46.1} & 43.3 & 43.0  \\ \hline
    Ours &  60.4   & \textbf{59.9} & \textbf{39.2} & \textbf{52.6} & 37.8 & \textbf{48.4} & \textbf{47.2} \\ 
                \hline                
			\end{tabular}
		}
	\end{center}
	\vspace{-1.5em}
	\caption{mAP@0.5 and $F_{mAP}$ scores on VOC2007 dataset.}
 \vspace{-0.5em}
	\label{tab:results_voc2007}
\end{table}

Results show that vanilla Fine-tuning behaves the worst as it pushes for the complete plasticity of the neural network. This is an extreme case in which the mAP for past classes is equal to zero or nearly so, which causes the $F_{mAP}$ score to be zero as we can see in the case of $B=10, I=10$.

LwF \cite{li2016_lwf} is a method that was initially proposed to tackle continual learning in classification task. It consists in distilling the current model from the previous one, in order to mimic its predictions. Even though this method provides good results in rehearsal-free continual classification \cite{belouadah2020_study}, it fails to generalize on the detection task. One reason could be that classification task is orders of magnitude  easier than object detection, and more specialized techniques are needed in order to tackle the latter.

 Our method outperforms results from state of the art in 11 cases out of 12. For VOC2007 dataset, in the scenario $B=19, I=1$, our method gains up to 14.4 mAP points compared to the state-of-the-art method SID \cite{peng2020_sid}. This is intuitive insofar as fixed-representations are more powerful when trained with more classes in the initial state. The diversity of the initial dataset and its size is crucial to have universal representation and transfer to unseen classes.

In the scenario $B=15, I=5$, $MultIOD$ outperforms SID in terms of mAP score, but fails to do so for the $F_{mAP}$ score. This indicates that in this specific case, SID balances better the compromise between plasticity (the ability to adapt to new class distribution), and stability (the ability to keep as much knowledge as possible from the past).

\begin{table}[htb!]
\setlength{\tabcolsep}{3pt}
	\begin{center}
		\resizebox{0.5\textwidth}{!}{
			\begin{tabular}{|l|c|c|c|c|c|c|c|c|}
				\hline
    Method & Full & \multicolumn{2}{c|}{$B=19$, $I=1$} & \multicolumn{2}{c|}{$B=15$, $I=5$} & \multicolumn{2}{c|}{$B=10$, $I=10$} \\	
				\hline
    &  $mAP$  & $mAP$ & $F_{mAP}$ & $mAP$ & $F_{mAP}$ & $mAP$ & $F_{mAP}$ \\ \hline
     
   	FT &  \multirow{6}{*}{73.1}  & 20.2 & 26.3 & 16.3 & 17.3 & 28.0 & 0.0  \\		
    LwF \cite{li2016_lwf} &  & 36.3 & 28.1 & 12.0 & 11.5 & 22.5  &  0.0\\	
    SDR 1 \cite{michieli2021_sdr} &   & 60.4 & 22.8 & 41.4 & 23.9 & 35.5 & 32.2\\
    SDR 2 \cite{michieli2021_sdr} &   & 49.8 & 25.8 & 21.9 & 22.3 & 30.1 &9.6\\
    SID \cite{peng2020_sid} &   & 41.6 & 19.3 & 48.4 & 17.9 & 45.5 & 44.2 \\ 
    GT' \cite{gang2022_pdm} &  & 65.2 & 39.2 &  54.0 &  33.8  & 51.2 & 50.9 \\ \hline
    Ours &  69.5 & \textbf{68.0} & \textbf{56.9} & \textbf{60.7} &  \textbf{47.0} & \textbf{56.6} &  \textbf{55.8} \\
                \hline                
			\end{tabular}
		}
	\end{center}
	\vspace{-1.5em}
	\caption{mAP@0.5 and $F_{mAP}$ scores on VOC0712 dataset.}
 \vspace{-1em}
	\label{tab:results_voc0712}
\end{table}

For VOC0712, $MultIOD$ outperforms other methods in all cases, followed by GT' \cite{gang2022_pdm}. It is noteworthy to mention that the experimental protocol of this method is not realistic insofar as authors remove images that contain objects from past classes when training the model on new classes. Thus, they avoid the problem of background shift by removing the overlap between annotations of background and  past class objects. However, in real-life situations, it is impossible to have this separation as both past and new class objects can appear in the scene, and the model should be able to effectively learn new classes while past ones are also there. SDR \cite{michieli2021_sdr} is a method that was initially proposed for continual semantic segmentation, and was later adapted by \cite{gang2022_pdm} to object detection. Regardless of the effectiveness of this method on semantic segmentation, it does not achieve the same impressive performance on continual object detection. It reaches good scores in the protocol $B=19, I=1$ only. One reason for that could be that in CenterNet there is less information to distill compared to segmentation where all object pixels are used.

Coincidentally, all CenterNet-based continual object detectors to which we compare $MultIOD$, use distillation to update model weights from its past states. Overall, results show the usefulness of transfer learning compared to distillation. This is true especially when classes learned incrementally belong to the same domain as classes learned initially.

In both Tables \ref{tab:results_voc2007} and \ref{tab:results_voc0712}, it is crucial to emphasize that our upper-bound (Full) model scores lower by 5.2 and 3.6 mAP points respectively, compared to the classical training  from the state of the art. Nevertheless,  it is interesting to notice that even with this drop in performance under traditional training, our model excels in comparison to other methods across nearly all scenarios involving incremental learning. This finding proves that $MultIOD$ is more robust against catastrophic forgetting, because the gap with our upper-bound model is tighter compared to the other methods with their respective upper-bound model. 

We remind that the difference in performance between the two upper-bound models is that our detector is built on top of  EfficientNet-B3 backbone, while the other methods use a ResNet50. In addition to the reduction in the number of parameters (17.7M vs. 32M), $MultIOD$  does not require keeping the past model to learn  new classes. In contrast, all methods used for comparison need the past  model in order to extract its activation and distill its knowledge to the current model. Thus, these methods require keeping two models simultaneously to make the training happen. Some predictions with $MultIOD$ are in Appendix \textcolor{red}{15}.

\subsection{Additional Experiment}
\label{subsec:additional}

In this experiment, we take the challenge to a more difficult level by adding only one or two classes in each incremental state. In this case, we use one feature pyramid for each class to have a complete separation of their representations in the upsampling and detection heads. To avoid an explosion in the number of parameters compared to the previous protocol, we reduce the number of filters in the feature pyramids to have a comparable number of parameters with the model used in previous experiments. Details of this architecture are in  Appendix \textcolor{red}{10}. 
Results in Table \ref{tab:effnetb3_protocol2_voc2007} show that we  lose from 1 to 6 mAP points approximately when passing from one state to another. Unfortunately, CenterNet-based state-of-the-art methods do not use this protocol, and we thus cannot compare them with $MultIOD$. We provide results for future use. 

\begin{table}[htb!]
	\begin{center}
		\resizebox{0.45\textwidth}{!}{
			\begin{tabular}{|l|c|c|c|c|c|c|}
				\hline

	  States & $S_0$ &  $S_1$ &  $S_2$ &  $S_3$ &  $S_4$ &  $S_5$ \\	\hline
     & $B=15$ & \multicolumn{5}{c|}{$I=1$} \\ 
				\hline
     FT    & \multirow{2}{*}{56.6}  & 0.7 & 0.3 & 1.0 & 0.3 & 1.8 \\
     Ours &&    53.4  & 47.9 & 45.6 & 43.3  & 42.7 \\
    \hline
      & $B=10$ & \multicolumn{5}{c|}{$I=2$} \\ 
				\hline
     FT  & \multirow{2}{*}{57.7} & 5.3 & 4.7 & 5.1 & 2.1 & 3.2  \\
     Ours &&   48.4    & 45.7 & 43.0 & 39.7  & 38.0 \\
    \hline
			\end{tabular}
		}
	\end{center}
	\vspace{-1.5em}
	\caption{mAP@0.5 of $MultIOD$ on  VOC2007.}
    \vspace{-0.5em}
	\label{tab:effnetb3_protocol2_voc2007}
\end{table}

\vspace{-1em}

\section{Conclusions}
\label{sec:conclusions}

In this paper, we present $MultIOD$, a class-incremental object detector based on CenterNet \cite{zhou2019_centernet}. Our approach uses a multihead detection component, along with a frozen backbone. A multihead feature pyramid is also used to ensure a satisfying trade-off between stability and plasticity. Finally, it involves an efficient class-wise NMS that provides robustness to remove duplicate bounding boxes.  Results show the effectiveness of our approach against CenterNet-based methods on Pascal VOC datasets \cite{everingham2010_pascal_voc} in many incremental scenarios. However, $MultIOD$ has some limitations:


\hspace{-0.2cm} $\bullet$ \textbf{Scalability -} One of the main drawbacks of our method is its limited capacity to scale, since we add one detection head for each incremental class. It would be interesting to investigate more intelligent ways to regroup classes by their semantic similarities, and create new detection heads only when the new classes are significantly different from already existing ones. This can be also combined with a gating mechanism \cite{aljundi2017_expert} that automatically selects relevant head(s) to use during inference.

\hspace{-0.2cm} $\bullet$ \textbf{Quality of the fixed representation -} This drawback is common to all techniques that rely on transfer learning. If the initial network is trained on sufficiently rich data, the quality of the transfer is prominent, but inversely, if the fixed representation is poor, the learning of subsequent classes drastically drops in performance. A good perspective would be to build more universal representations \cite{wang2023_universal} and test the transferability of better datasets, such as COCO \cite{lin2014_coco}.

\hspace{-0.3cm} \textbf{Acknowledgment}: This work was granted access to the HPC resources of IDRIS under the allocation 2023-A0141014124 made by GENCI.


{
    \small
    \bibliographystyle{ieeenat_fullname}
    \bibliography{main}
}

\clearpage
\setcounter{page}{1}
\maketitlesupplementary

\section{Illustration of Background Interference}
\label{appendix:background_interference}

\begin{figure}[thb!]
	\begin{center}
    \includegraphics[width=0.49\textwidth]{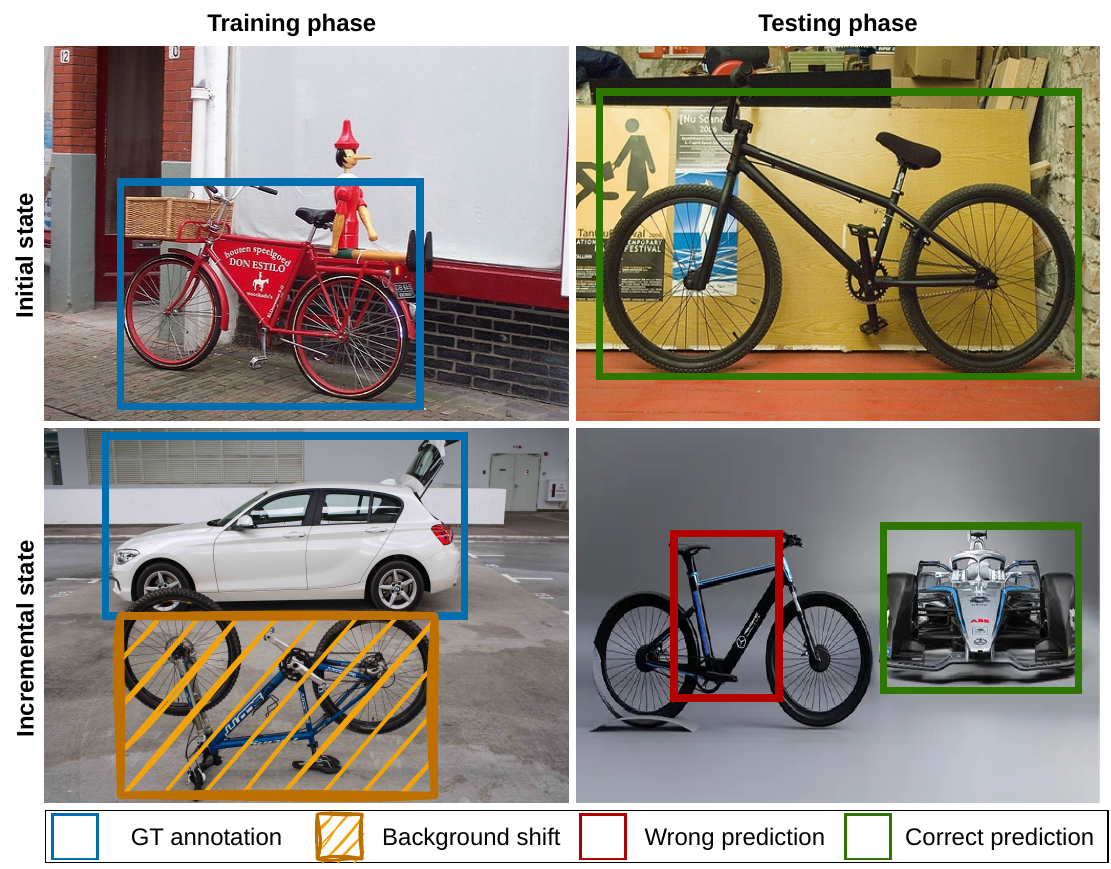}
    \vspace{-0.5cm}
	\end{center}
	\caption{ Illustration of background interference
 }
	\vspace{-1em}
	\label{fig:catastrophic_forgetting}
\end{figure}

Figure \ref{fig:catastrophic_forgetting} illustrates background interference.  In the initial state, the model learns correctly how to predict bicycles. In the incremental state, the bicycle is not anymore annotated, which causes: (1) background shift as the bicycle is confused with background, and (2) catastrophic forgetting because of distribution shift towards the new class. Here, the model learns correctly cars but fails to detect bicycles.

\section{Implementation Details}
\label{appendix:implementation_details}

We implement our method using Keras (Tensorflow 2.8.0). We use Adam optimizer with a learning rate of $2e^{-4}$, a batch size of $16$, and a weight decay of $ 1.25e-5$. Similarly to \cite{zhou2019_centernet}, we use images of size $512 \times 512$, down-sampled $4$ times to have prediction maps of size $128 \times 128$. All our models and those of compared methods are pretrained with imagenet weights. We use a detection threshold of 5\% to compute the mean-average-precision (mAP), even thought we could use lower threshold to improve results, we prefer to keep the model inference time bounded (in the state of the art, a value of 1\% is usually used). 

\begin{figure}[t]
	\begin{center}
    \includegraphics[width=0.5\textwidth]{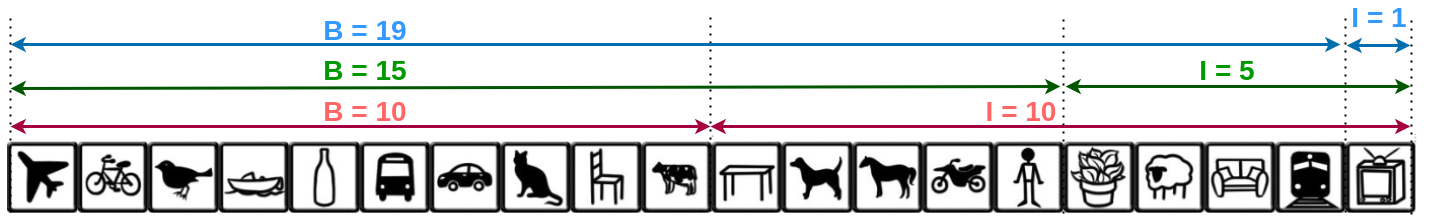}
	\end{center}
    \vspace{-1em}
	\caption{Pascal VOC incremental protocol}
	\label{fig:voc_incremental_protocol}
\end{figure}

For VOC datasets \cite{everingham2010_pascal_voc}, we train our model for $70$ epochs in each state, we decay the learning rate by $10$ at epochs $45$ and $60$. For training, we use as augmentations random flip, random resized crop, color jittering, and random scale. For testing, we use flip augmentation like in \cite{peng2020_sid, zhou2019_centernet, gang2022_pdm}. As mentioned in the main paper, the classes of VOC are ordered alphabetically before being divided into groups. Figure \ref{fig:voc_incremental_protocol} shows this order and reminds the protocol used.

For MNIST dataset \cite{lecun2010mnist}, we train our model for $20$ epochs in each state, and keep the other hyper-parameters unchanged. For training, we use as augmentations  random resized crop and random scale. For testing, we use flip augmentation.

\section{Feature Pyramids Architecture}
\label{appendix:feature_pyramids_architecture}

\begin{figure}[thb!]
	\begin{center}
    \includegraphics[width=0.4\textwidth]{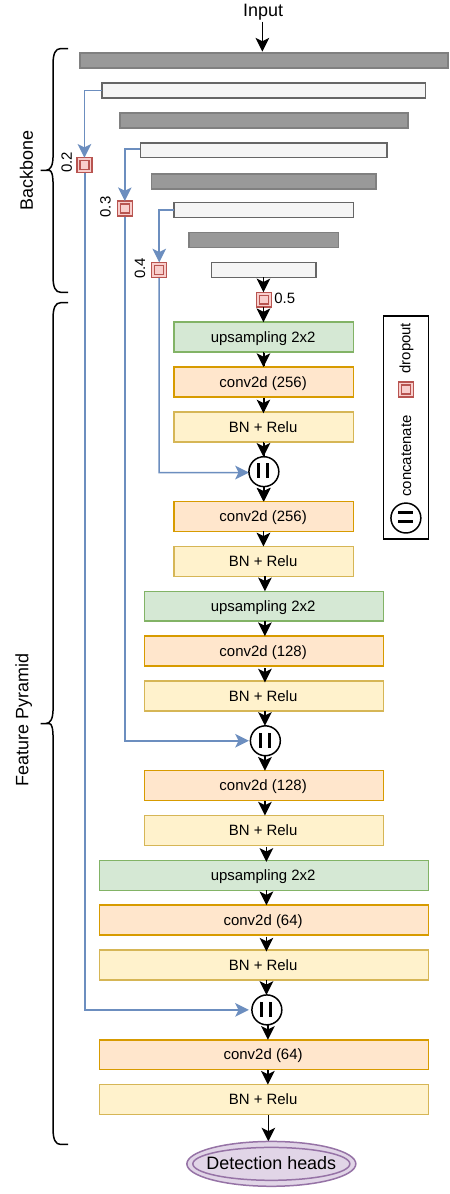}
    \vspace{-0.5cm}
	\end{center}
	\caption{Architecture of one Feature Pyramid in $MultIOD$
 }
	\vspace{-1em}
	\label{fig:fpn}
\end{figure}

In $MultIOD$, each feature pyramid  is constructed of 4 levels that are connected using dropout layers to the backbone. The connected layers of backbone are colored in light gray in Figure \ref{fig:fpn}, and are specified for each EfficientNet variant in Table \ref{tab:fpn_layers}. Layer names given in this table are based on the official $keras-applications$ implementations.

As shown in Figure \ref{fig:fpn}, each feature pyramid contains three blocks of layers each containing: upsampling $2\times2$, convolution layer with number of filters shown between parenthesis, batch normalization layer and ReLU, concatenation layer, another convolutional layer, batch norm and ReLU. Upsampling is done progressively in order to capture multi-scale features. We use the FPN implementation of this GitHub repository \footnote{https://github.com/Ximilar-com/xcenternet}. In the class-wise feature pyramids (Subsection \textcolor{red}{5.2} of the main paper), we use the same architecture described in Figure \ref{fig:fpn}, but we reduce the number of filters in the convolutional layers to avoid an explosion in the number of parameters. We thus use only 64, 64, 32, 32, 16, and 16 filters in each convolutional layer, respectively.

\section{MNIST Dataset Creation Details}
\label{appendix:mnist_dataset}

We made sure to create a challenging dataset by doing the following:
\begin{itemize}
    \item We set the minimum and maximum digit sizes between 50²  and 200² pixels, respectively, in order to have both small and large digits.
    \item We make sure to have one to five digits in each image, for diversification.
    \item The background shift is present in this dataset as we randomly pick digits from a set of ten, regardless of the current state.
\end{itemize}

\begin{figure*}[htb!]
    \centering
    \begin{subfigure}[b]{0.19\textwidth}
        \includegraphics[width=\textwidth]{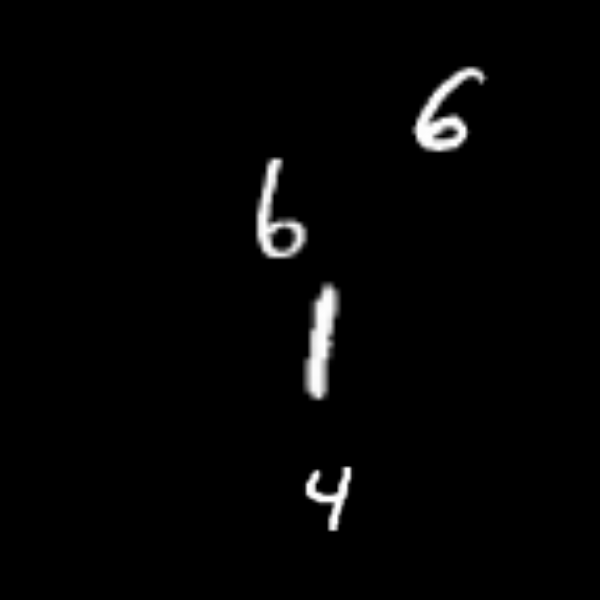}
    \end{subfigure}
    \begin{subfigure}[b]{0.19\textwidth}
        \includegraphics[width=\textwidth]{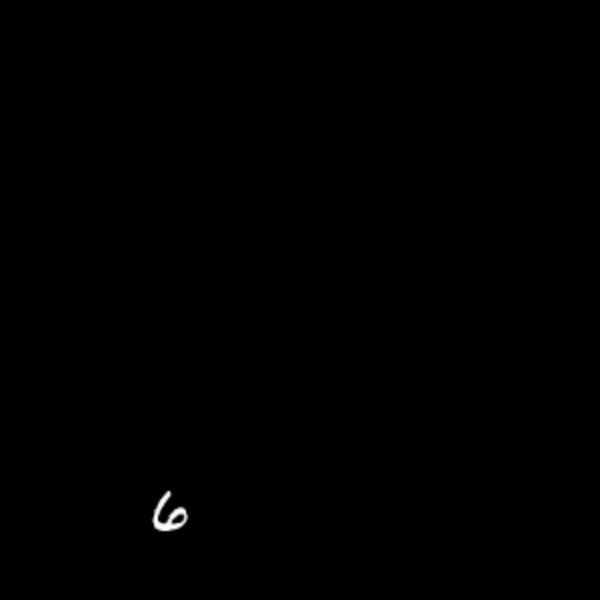}
    \end{subfigure}
    \begin{subfigure}[b]{0.19\textwidth}
        \includegraphics[width=\textwidth]{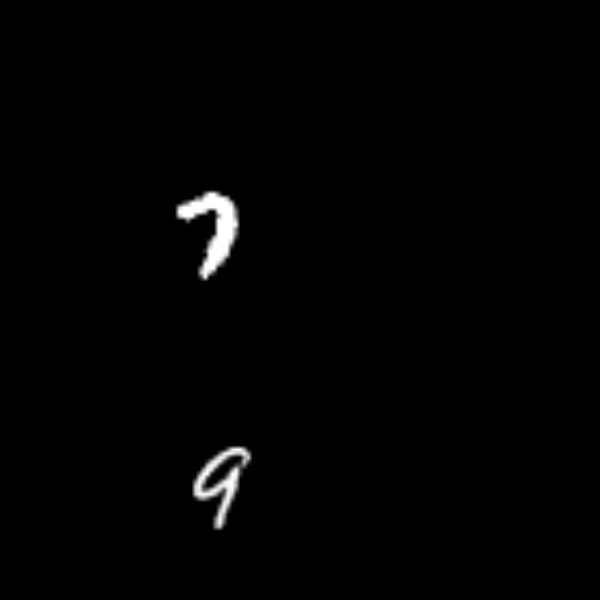}
    \end{subfigure}
    \begin{subfigure}[b]{0.19\textwidth}
        \includegraphics[width=\textwidth]{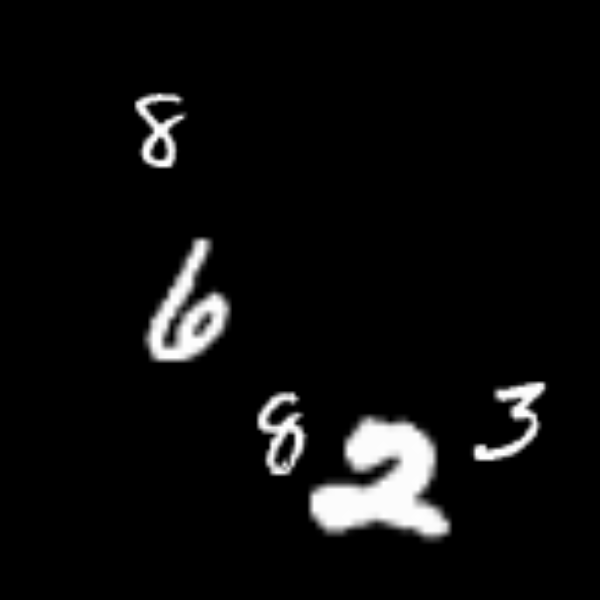}
    \end{subfigure}
    \begin{subfigure}[b]{0.19\textwidth}
        \includegraphics[width=\textwidth]{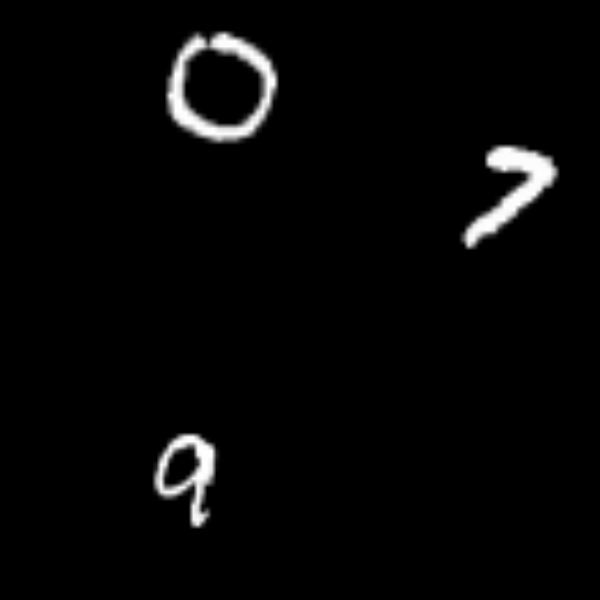}
    \end{subfigure}
    \begin{subfigure}[b]{0.19\textwidth}
        \includegraphics[width=\textwidth]{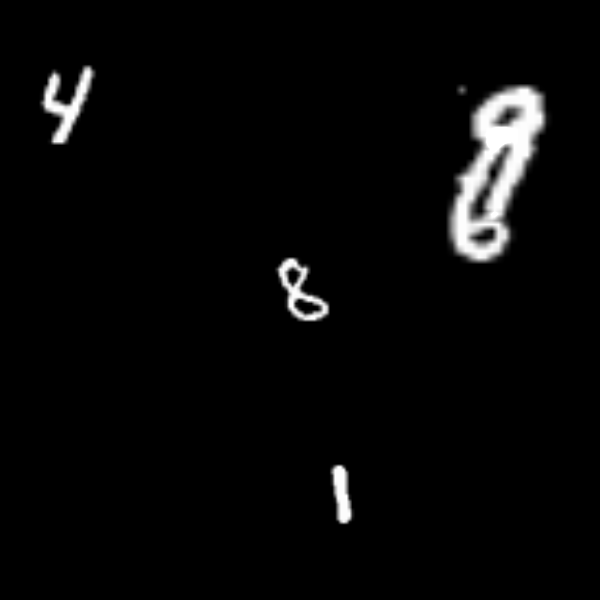}
    \end{subfigure}
    \begin{subfigure}[b]{0.19\textwidth}
        \includegraphics[width=\textwidth]{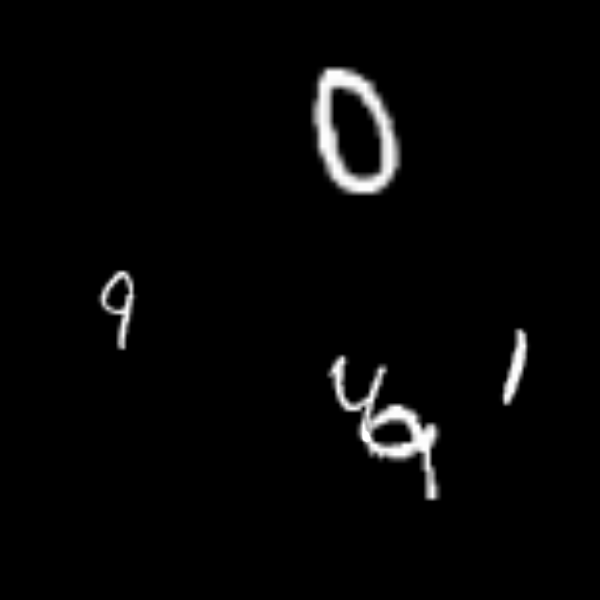}
    \end{subfigure}
    \begin{subfigure}[b]{0.19\textwidth}
        \includegraphics[width=\textwidth]{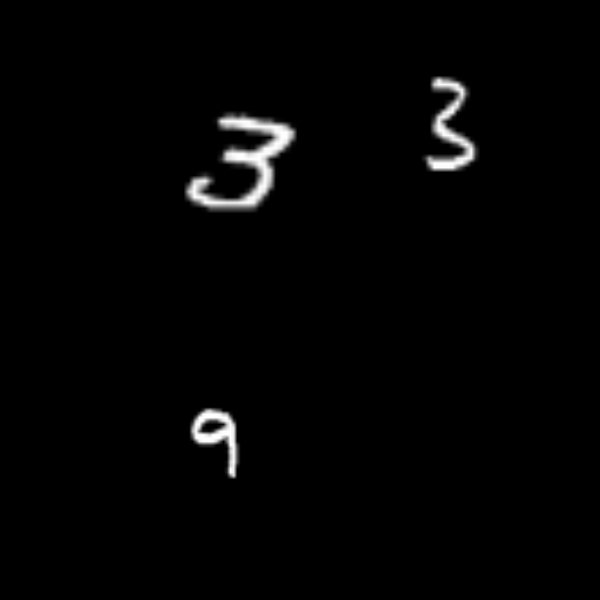}
    \end{subfigure}
    \begin{subfigure}[b]{0.19\textwidth}
        \includegraphics[width=\textwidth]{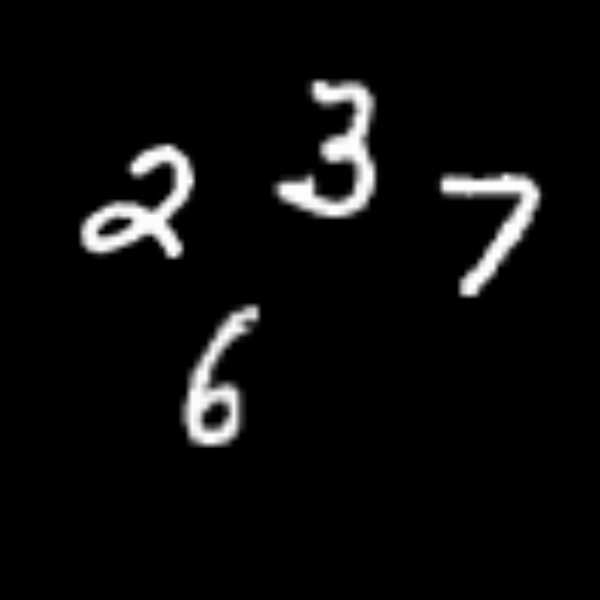}
    \end{subfigure}
    \begin{subfigure}[b]{0.19\textwidth}
        \includegraphics[width=\textwidth]{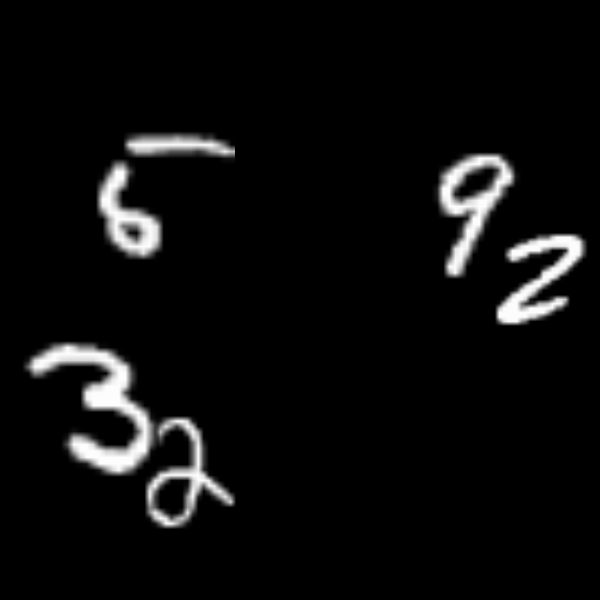}
    \end{subfigure}
    \caption{Examples of generated MNIST images}
    \label{fig:mnist_images}
\end{figure*}

Examples of generated images are in Figure \ref{fig:mnist_images}.

\begin{table*}[htb!]
	\begin{center}
		\resizebox{0.8\textwidth}{!}{
			\begin{tabular}{|l|c|c|c|c|}
				\hline
    Backbone  & Level 1 & Level 2 & Level 3 & Level 4 \\	
				\hline
    
    EfficientNet-B0 &   block2b-activation & block3b-activation & block5c-activation & top-activation \\
    EfficientNet-B3 &   block2c-activation & block3c-activation & block5e-activation & top-activation \\
    EfficientNet-B5 &   block2e-activation & block3e-activation & block5g-activation & top-activation \\
    \hline     
			\end{tabular}
		}
	\end{center}
	\vspace{-1.5em}
	\caption{Names of layers in Keras corresponding to Feature Pyramid \cite{lin2016_fpn} Levels for different EfficientNet architectures}
	\label{tab:fpn_layers}
\end{table*}

\section{Results with mAP@[0.5, 0.95]}
\label{appendix:results_of_95}

Tables \ref{tab:supp_results_voc2007} and \ref{tab:supp_results_voc0712} provide results of our method on VOC2007 and VOC0712, using mAP averaged over IoU threshold that varies between 0.5 and 0.95 with a step of 0.05. Results are provided for future comparisons.

\begin{table}[htb]
\setlength{\tabcolsep}{3pt}
	\begin{center}
		\resizebox{0.5\textwidth}{!}{
			\begin{tabular}{|l|c|c|c|c|c|c|c|c|}
				\hline
	Method & Full & \multicolumn{2}{c|}{$B=19$, $I=1$} & \multicolumn{2}{c|}{$B=15$, $I=5$} & \multicolumn{2}{c|}{$B=10$, $I=10$} \\	
				\hline
     & $mAP$  & $mAP$ & $F_{mAP}$ & $mAP$ & $F_{mAP}$ & $mAP$ & $F_{mAP}$ \\ \hline
    IoU = 0.5 &  60.4   & 59.9 & 39.2 & 52.6 & 37.8 & 48.4 & 47.2 \\ 
    IoU = [0.5, 0.95] &  35.9   & 35.7 & 18.5 & 30.7 & 20.2 & 25.9 & 23.9 \\ 
                \hline                
			\end{tabular}
		}
	\end{center}
	\vspace{-1.5em}
	\caption{Mean-average-precision and $F_{mAP}$ score on VOC2007.}
	\label{tab:supp_results_voc2007}
\end{table}

\begin{table}[htb]
\setlength{\tabcolsep}{3pt}
	\begin{center}
		\resizebox{0.5\textwidth}{!}{
			\begin{tabular}{|l|c|c|c|c|c|c|c|c|}
				\hline
    Method & Full & \multicolumn{2}{c|}{$B=19$, $I=1$} & \multicolumn{2}{c|}{$B=15$, $I=5$} & \multicolumn{2}{c|}{$B=10$, $I=10$} \\	
				\hline
    &  $mAP$ & $mAP$ & $F_{mAP}$ & $mAP$ & $F_{mAP}$ & $mAP$ & $F_{mAP}$ \\ \hline
     
    IoU = 0.5 &  69.5 & 68.0 & 56.9 & 60.7 &  47.0 & 56.6 & 55.8 \\
     IoU = [0.5, 0.95] &  45.7  & 44.5 & 33.6 & 39.0 & 26.8 & 33.2 & 31.0 \\ 
                \hline                
			\end{tabular}
		}
	\end{center}
	\vspace{-1.5em}
	\caption{Mean-average-precision and $F_{mAP}$ score on VOC0712.}
	\label{tab:supp_results_voc0712}
\end{table}

\section{Ablation of Backbones on MNIST}
\label{appendix:ablation_backbones}

In Table \ref{tab:ablation_backbone_mnist}, we provide results of $MultIOD$ using different backbones on MNIST dataset. Because this dataset is not challenging and is of a small size, it is easier for large models like EfficientNet \cite{tan2019_effnet} to learn it. In our experiments, it is hard to determine which backbone provides the best results for this dataset, as each backbone is best in one configuration. However, results of different models are comparable, and we thus recommend using the smallest version (EfficientNet-B0) for this dataset.

\begin{table}[htb!]
\setlength{\tabcolsep}{3pt}
	\begin{center}
		\resizebox{0.5\textwidth}{!}{
			\begin{tabular}{|l|c|c|c|c|c|c|c|c|}
				\hline
    Method & Full & \multicolumn{2}{c|}{$B=9$, $I=1$} & \multicolumn{2}{c|}{$B=7$, $I=3$} & \multicolumn{2}{c|}{$B=5$, $I=5$} \\	
				\hline
    & $mAP$  & $mAP$ & $F_{mAP}$ & $mAP$ & $F_{mAP}$ & $mAP$ & $F_{mAP}$ \\ \hline
     
    EfficientNet-B0 &  93.1 & 91.3  & 91.2 &  \textbf{93.1} &  \textbf{93.5} & 91.3 &  91.3\\
    EfficientNet-B3 &  91.1 & \textbf{91.7}  & \textbf{92.6} &  92.0 &  92.4 & 89.7 &  89.7 \\
    EfficientNet-B5  &   \textbf{93.7} & 91.2  & 92.4 &  90.6 &  91.4 & \textbf{92.5} &  \textbf{92.5} \\
                \hline                
			\end{tabular}
		}
	\end{center}
	\vspace{-1.5em}
	\caption{Ablation of backbones on MNIST dataset (mAP@0.5).}
	\label{tab:ablation_backbone_mnist}
\end{table}

\section{Ablation of NMS Strategies on VOC2007}
\label{appendix:ablation_nms}

In Table \ref{tab:ablation_nms_voc}, we provide the results of $MultIOD$ using different NMS strategies on VOC2007 dataset. Similarly to the results presented in the main paper, the method that achieves the best results is class-wise NMS, followed by inter-class NMS. Soft-NMS and No-NMS are the methods that achieve the lowest results.

\begin{table}[htb!]
	\begin{center}
		\resizebox{0.5\textwidth}{!}{
			\begin{tabular}{|l|c|c|c|c|c|c|c|}
				\hline
    Method  & Full & \multicolumn{2}{c|}{$B=19$, $I=1$} & \multicolumn{2}{c|}{$B=15$, $I=5$} & \multicolumn{2}{c|}{$B=10$, $I=10$} \\	
				\hline
    
    & $mAP$  & $mAP$ & $F_{mAP}$ & $mAP$ & $F_{mAP}$ & $mAP$ & $F_{mAP}$ \\ \hline
    No-NMS &    51.7 & 51.6 & 33.3 & 44.4 & 28.7 & 36.9 &  33.2 \\ 
    Soft-NMS  &      45.8 & 46.6 & 29.6 &  40.5 &  23.8 & 34.5 &  31.4 \\ 
    Inter-class NMS &     53.0 & 51.8 &  35.7 & 46.1 & 34.1 & 41.9 &  40.1 \\
    Class-wise NMS &    \textbf{56.7} & \textbf{55.7} & \textbf{35.6} &  \textbf{49.2} &  \textbf{33.8} & \textbf{46.3} &  \textbf{45.5} \\ \hline   
     
			\end{tabular}
		}
	\end{center}
	\vspace{-1.5em}
	\caption{Performance of our model using  VOC2007 dataset with different NMS strategies and EfficientNet-B0.}
	\label{tab:ablation_nms_voc}
\end{table}

\section{Examples of Detections with MultIOD}
\label{appendix:examples_detection}

\begin{figure*}[ht!]
    \centering
    \begin{subfigure}[b]{0.32\textwidth}
        \includegraphics[width=\textwidth]{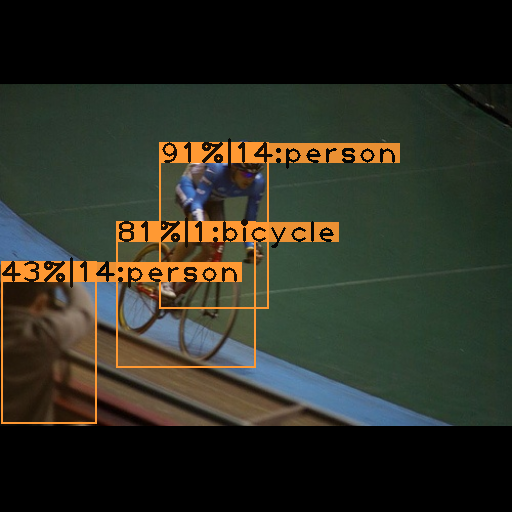}
    \end{subfigure}
    \begin{subfigure}[b]{0.32\textwidth}
        \includegraphics[width=\textwidth]{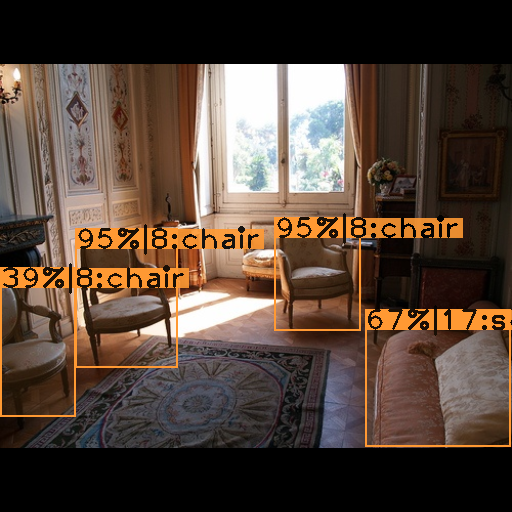}
    \end{subfigure}
    \begin{subfigure}[b]{0.32\textwidth}
        \includegraphics[width=\textwidth]{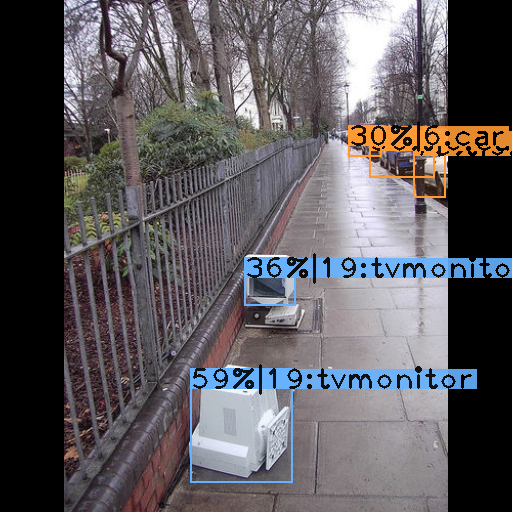}
    \end{subfigure}
    \begin{subfigure}[b]{0.32\textwidth}
        \includegraphics[width=\textwidth]{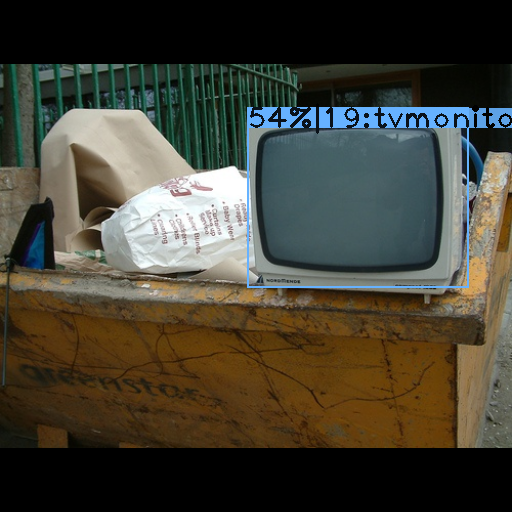}
    \end{subfigure}
    \begin{subfigure}[b]{0.32\textwidth}
        \includegraphics[width=\textwidth]{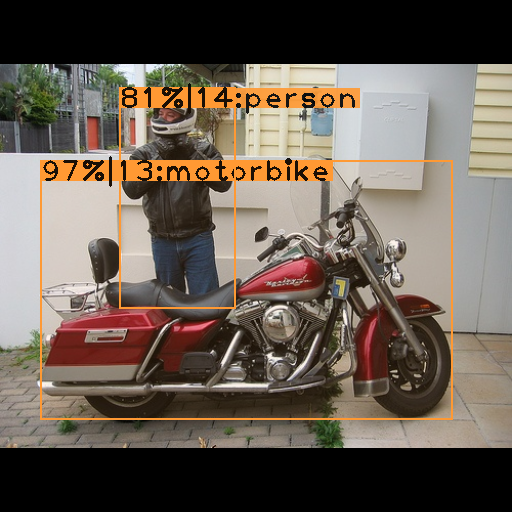}
    \end{subfigure}
    \begin{subfigure}[b]{0.32\textwidth}
        \includegraphics[width=\textwidth]{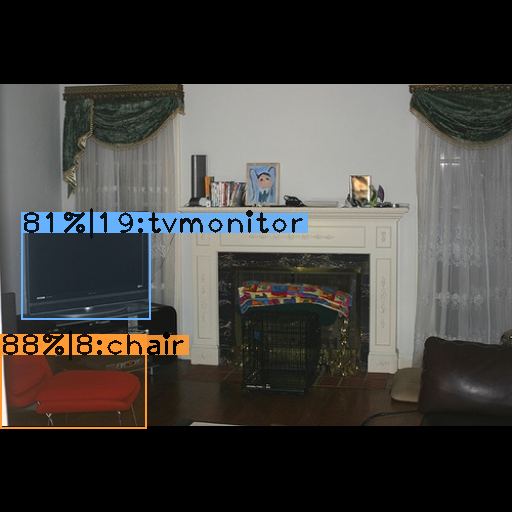}
    \end{subfigure}

        \begin{subfigure}[b]{0.32\textwidth}
        \includegraphics[width=\textwidth]{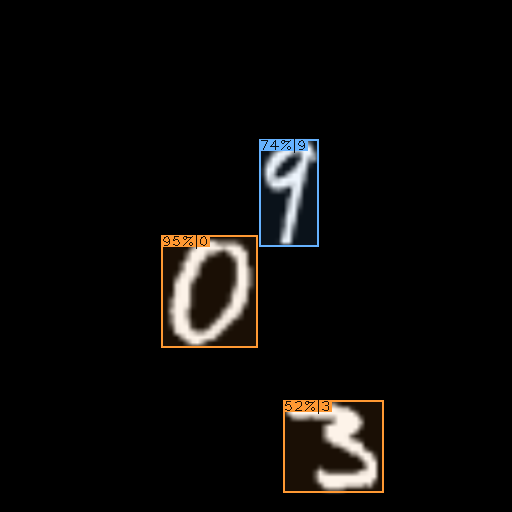}
    \end{subfigure}
    \begin{subfigure}[b]{0.32\textwidth}
        \includegraphics[width=\textwidth]{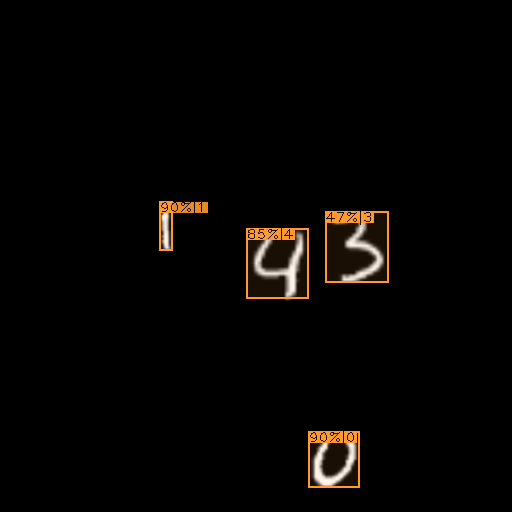}
    \end{subfigure}
    \begin{subfigure}[b]{0.32\textwidth}
        \includegraphics[width=\textwidth]{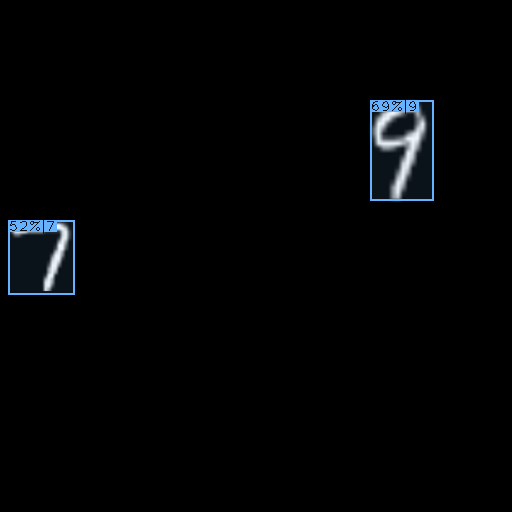}
    \end{subfigure}
    \begin{subfigure}[b]{0.32\textwidth}
        \includegraphics[width=\textwidth]{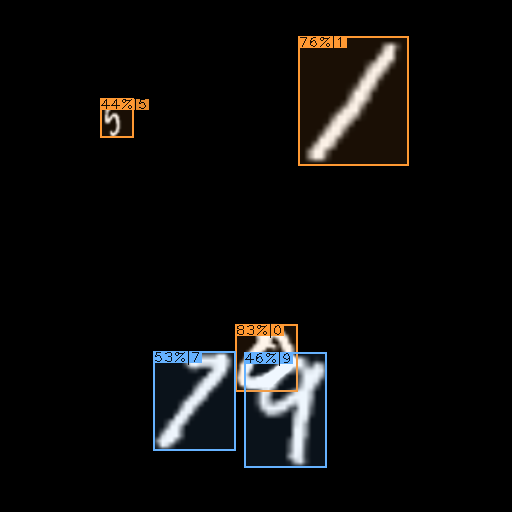}
    \end{subfigure}
    \begin{subfigure}[b]{0.32\textwidth}
        \includegraphics[width=\textwidth]{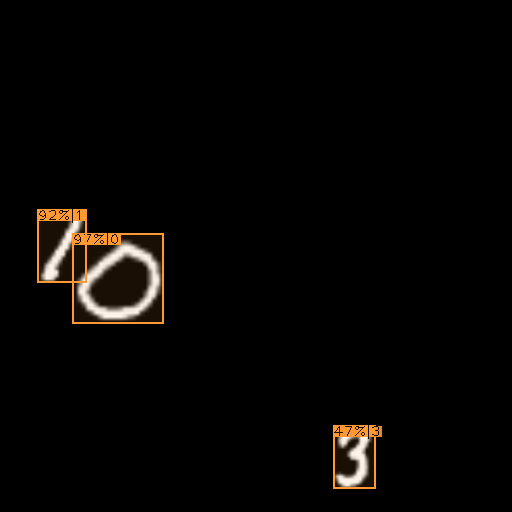}
    \end{subfigure}
    \begin{subfigure}[b]{0.32\textwidth}
        \includegraphics[width=\textwidth]{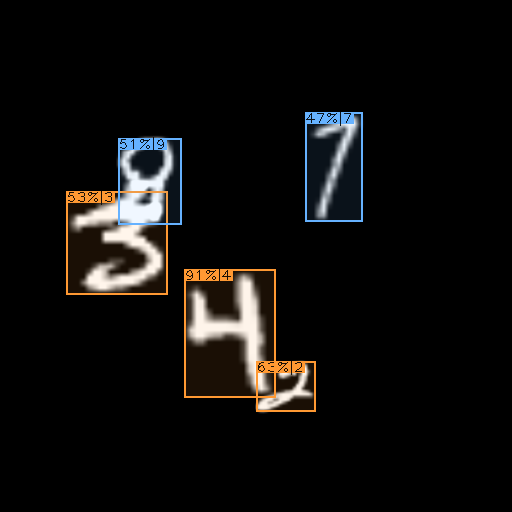}
    \end{subfigure}

    \caption{Examples of detections with $MultIOD$ on VOC0712 (EfficientNet-B3, B=19, I=1) and MNIST (EfficientNet-B0, B=7, I=3)}
    \label{fig:image_results}
\end{figure*}

Figure \ref{fig:image_results} provides examples of predictions made with our $MultIOD$ continual  detector. Orange is used for past class detections, and blue is used for new class detections. Visual results confirm the robustness of our method against catastrophic forgetting. $MultIOD$ provides a good compromise between stability of the neural network and its plasticity.

\section{Comparison Against Two-Stage Detectors}
\label{appendix:comparison_rehearsal_based}

Table \ref{fig:image_results} provides a comparison of $MultIOD$ with some two-stage continual detectors on VOC2007 dataset. Rehearsal-based methods store a subset of past data, and replay it when training new classes to tackle catastrophic forgetting. 

\begin{table}[htb!]
\setlength{\tabcolsep}{3pt}
	\begin{center}
		\resizebox{0.5\textwidth}{!}{
			\begin{tabular}{|l|c|c|c|c|c|}
				\hline
    Method  & Detector & Rehearsal? & \multicolumn{1}{c|}{$B=19$, $I=1$} & \multicolumn{1}{c|}{$B=15$, $I=5$} & \multicolumn{1}{c|}{$B=10$, $I=10$} \\	
				\hline
    MultIOD  & CenterNet & $\times$ & 59.9 & 52.6 & 48.4  \\
    MVD \cite{dongbao2022_multi}  & Faster R-CNN & $\times$ & 69.7 & 66.5 & 66.1  \\ 
    IncDet \cite{liu2021_incdet}  & Fast(er) R-CNN & $\times$ & $\times$ & 70.4 & 70.8 \\ 
    RD-IOD \cite{dongbao2022_rd_iod} & Faster R-CNN & $\times$ & 72.1 & 69.7 & 66.2  \\ 
    Faster-ILOD \cite{peng2020_faster_ilod} & Faster R-CNN & $\times$ & 68.6 & 68.0 & 62.2  \\ 
    ORE \cite{joseph2021_ore}  & Faster R-CNN & $\checkmark$ & 68.9 & 68.5 & 64.6  \\ 
    OST \cite{yang2022_semantic}  & Faster R-CNN & $\checkmark$ & 69.8 & 69.9 & 65.0  \\   
    \hline                
	\end{tabular}
		}
	\end{center}
	\vspace{-1.5em}
	\caption{mAP@0.5 scores on VOC2007 dataset.}
 \vspace{-1em}
	\label{tab:supp_results_rehearsal}
\end{table}

Results indicate that $MultIOD$ achieves the lowest results compared to methods that combine both two-stage detectors and rehearsal memory. This is intuitive because with the absence of memory from the past, inter-class separability becomes more challenging.

Fast(er)-RCNN are two-stage detectors that perform better than CenterNet, but are very slow which make them not suitable for real-life applications. A trade-off is required to select between the two detectors depending on the use case.

%


\end{document}